\pdfoutput=1

\documentclass[11pt]{article}

\usepackage{acl}

\usepackage{times}
\usepackage{latexsym}

\usepackage[T1]{fontenc}

\usepackage[utf8]{inputenc}

\usepackage{microtype}

\usepackage{inconsolata}

\usepackage{graphicx}
\usepackage{colortbl}
\usepackage{booktabs}
\usepackage{verbatim}
\usepackage{ntheorem}
\theoremseparator{:}
\newtheorem{hyp}{Hypothesis}

\title{Taking Action Towards Graceful Interaction: The Effects of Performing Actions on Modelling Policies for Instruction Clarification Requests}

\author{Brielen Madureira$^{\mathbf{1}}$ \hspace{10mm}  David Schlangen$^{\mathbf{1, 2}}$ \\
$^{\mathbf{1}}$Computational Linguistics, Department of Linguistics \\ University of Potsdam, Germany \\
$^{\mathbf{2}}$German Research Center for Artificial Intelligence (DFKI), Berlin, Germany \\
\texttt{\{madureiralasota,david.schlangen\}@uni-potsdam.de}}

\begin{document}
\maketitle

\begin{abstract}
Clarification requests are a mechanism to help solve communication problems, \textit{e.g.}~due to ambiguity or underspecification, in instruction-following interactions. Despite their importance, even skilful models struggle with producing or interpreting such repair acts. In this work, we 
test three hypotheses concerning the effects of action taking as an auxiliary task in modelling iCR policies. Contrary to initial expectations, we conclude that its contribution to learning an iCR policy is limited, but some information can still be extracted from prediction uncertainty. We present further evidence that even well-motivated, Transformer-based models fail to learn good policies for \textit{when to ask} Instruction CRs (iCRs), while the task of determining \textit{what to ask about} can be more successfully modelled. Considering the implications of these findings, we further discuss the shortcomings of the data-driven paradigm for learning meta-communication acts.
\end{abstract}

\section{Introduction}
\label{sec:intro}

The concept of \textit{graceful interaction} \citep{hayes1979graceful,hayes1983steps} was proposed as a set of skills that machines should exhibit to properly engage in cooperative dialogue with humans, among which are being able to ask for, understand and offer clarification. More than forty years later, the ineptitude of large language models and voice assistants to handle underspecifications and to properly process or produce clarification requests (CR) is still being documented \citep{larsson-2017-user,kuhn2022clam,li-etal-2023-python,deng2023prompting,shaikh2023grounding}. It is also one of the acknowledged limitations of the currently prevailing commercial chat-optimised LLM.\footnote{In the blogpost releasing chatGPT, the limitations section says: ``\textit{Ideally, the model would ask clarifying questions when the user provided an ambiguous query. Instead, our current models usually guess what the user intended.}''. Source: \url{https://openai.com/blog/chatgpt}.}

\begin{figure}[ht!]
    \centering
    \includegraphics[trim={0 7cm 9.5cm 0},clip,width=0.85\linewidth]{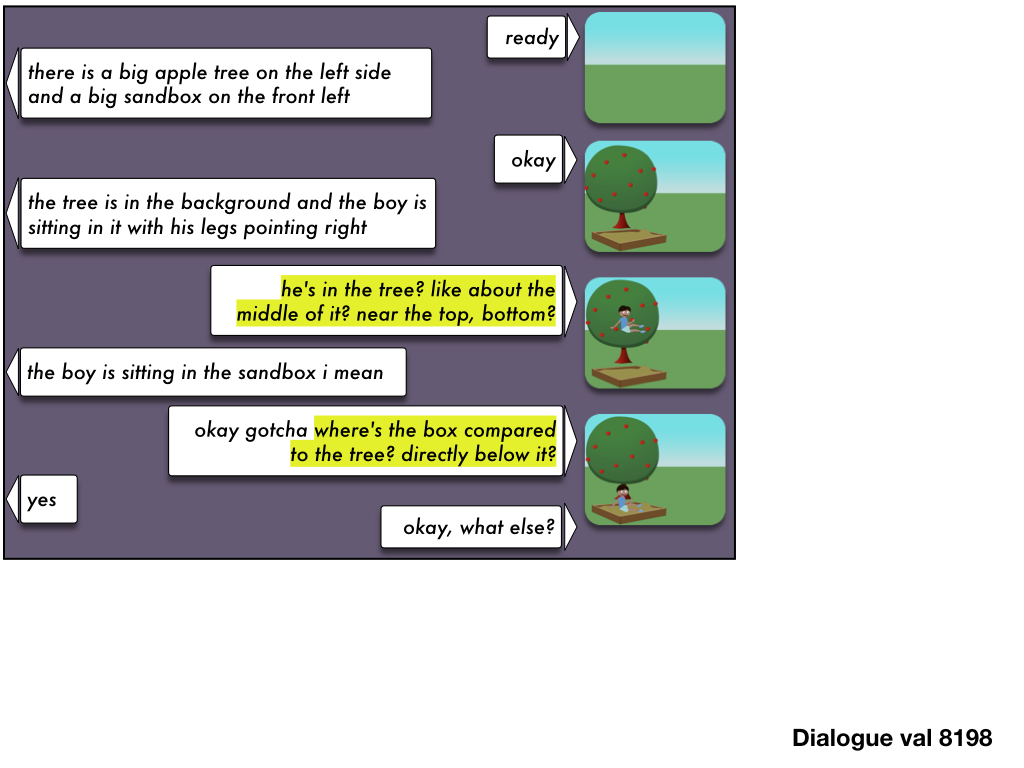}
    \caption{Clarification requests posed by an instruction follower, 
    demonstrating uncertainty on deciding what actions to take due to ambiguity or underspecification. From: CoDraw dialogue game 8198, \href{https://creativecommons.org/licenses/by-nc/4.0/}{CC BY-NC 4.0}, cliparts from \citet{zitnick2013bringing}.}
    \label{fig:example-icr}
\end{figure}

Given that they are modulated for instructions, this seems to be a peculiar fault: CRs are a crucial mechanism used to repair misunderstandings in instruction following interactions \citep{benotti-2009-clarification}, as we see in Figure \ref{fig:example-icr}. On second thought, it comes as no surprise. Clarification exchanges are meta-communication acts that do not normally appear in non-interactive data \citep{kuhn2022clam} and are also relatively rare in dialogue data. As a specific dialogue phenomenon, CRs have an empirical frequency of 4\% of turns in spontaneous conversations to 11\% of turns in strictly instruction-following interactions \citep{purver-etal-2001-means,benotti-blackburn-2021-recipe,madureira-schlangen-2023-instruction}. Therefore, it is still unclear to what extent CR strategies can be learnt with data-driven approaches \citep{benotti-blackburn-2021-recipe}.

Many existing CR datasets, despite their utility for applications like conversational search \citep{keyvan2022approach,rahmani-etal-2023-survey}, either have not been collected via real interactions or are synthetic, so that learnt CR policies may not correspond to genuine human behaviour. Moreover, current best-performing data-driven models are still not doing very well in deciding when to request clarification (see \S\ref{sec:litreview}), and we must understand why.

CRs can occur in all four levels of communication \citep{clark1996using}: Attention (due to problems in the channel), identification (due to acoustic impediments), recognition (when the signal is understood but a lexical, parsing or reference problem manifests) and consideration (when the intention is unclear) \citep{rodriguez2004form}. Instruction CRs (iCR) emerge mostly at Clark's 4th level of communication \citep{clark1996using}, \textit{i.e.}~at the level of uptake \citep{schloder-fernandez-2014-clarification}, to solve ambiguities and underspecifications. 

Recently, \citet{madureira-schlangen-2023-instruction,madureira2023are} have argued that the multimodal CoDraw game \citep{kim-etal-2019-codraw} is a rich resource for iCRs, naturally produced as a by-product of game playing via actions, as in the example in Figure \ref{fig:example-icr}. This dataset offers a balance between size (in comparison to well-curated but small corpora) and retaining ecological validity (as opposed to massive datasets collected or crafted artificially). Supposing underlying iCR strategies can emerge from data, we can reasonably assume that action-taking is a key component in modelling policies for deciding when and what to repair in this type of game. 

However, one major drawback of the proposed baseline models is the \textit{overhearer} paradigm: Models are not trained to act as authentic dialogue participants. Instead, they process other people's interactions, and at some points have to predict when to ask iCRs, a decision detached from the actual actions required by the game. Understanding is different for overhearers and addressees, and the latter have advantages in building common ground \citep{schober1989understanding}. \citet{clark1992arenas} argues that subjects in psycholinguistics are actually usually treated as overhearers; we add to that that many NLP approaches are also modelling overhearers.

\paragraph{Contributions} Given that background, this work aims to expand the boundaries of the open question of learning meta-communication acts from human data. We do that by (i) implementing a more well-motivated model for learning \textit{when to ask} iCRs in CoDraw; (ii) taking another step towards a more realistic agent by defining and modelling the task of \textit{what to ask} about; and, most importantly, (iii) testing three hypotheses to study the effect of action-taking in learning iCR policies, verifying whether a measure of certainty can be used to probe for iCR abilities and inform predictions.

\section{Related Work}
\label{sec:litreview}
\paragraph{Learning \textit{when to ask} questions} The problem of knowing when to ask questions in an interaction appears in various contexts. Relevant work has been done in language-aided visual navigation \citep{nguyen-daume-iii-2019-help,thomason2020vision,chi2020just,nguyen2022framework}, in which the agent must take actions in an environment and decide when to ask for help, where RL is a suitable method. Similar policies are necessary in interactive settings like visual dialogue games that require deciding when to stop asking \citep{shekhar-etal-2018-ask} or incremental predictions on when to answer a question \citep{boyd-graber-etal-2012-besting}. 

\paragraph{Modelling clarification requests} A vast literature exists on describing and modelling clarification strategies \citep[\textit{inter alia}]{purver2003means,gabsdil2003clarification,schlangen-2004-causes,rodriguez2004form,rieser-lemon-2006-using,stoyanchev-etal-2013-modelling}. In the age of neural network-based NLP, the problem has commonly been broken down into various tasks that are learnt from data: \textit{When to ask} \citep{narayan-chen-etal-2019-collaborative,aliannejadi-etal-2021-building,shi-etal-2022-learning,kiseleva2022iglu}, \textit{what to ask} about \citep{braslavski2017cqa,aliannejadi-etal-2021-building,hu-etal-2020-interactive}, and how to generate \citep{kumar-black-2020-clarq,gervits-etal-2021-agents,majumder-etal-2021-ask} or select/rank appropriate CRs \citep{rao-daume-iii-2018-learning,aliannejadi2019asking,mohanty2023transforming}. Ideally, these tasks should be tied into a single agent, but several works are still approaching the problem in a ``task-framed'' fashion without integration of all capabilities \citep{schlangen-2021-targeting}.

Modelling policies for \textit{when to ask} for clarification in instruction following is far from being a solved problem, as models perform well below the ceiling. The performance in the Minecraft Dialogue dataset is 0.63 accuracy for the CR class \citep{shi-etal-2022-learning}. In the recent IGLU challenge \citep{kiseleva2022iglu}, the best model in the leaderboard\footnote{Reported in the \href{https://www.aicrowd.com/challenges/neurips-2022-iglu-challenge/problems/neurips-2022-iglu-challenge-nlp-task/leaderboards}{NeurIPS 2022 IGLU challenge platform.}} reaches 0.75 weighted average F1 Score. In predicting underspecification for code generation, the highest performance is 0.78 binary F1Score \citep{li-etal-2023-python}.
In Codraw-iCR, the baseline achieves a similarly suboptimal 0.34 average precision \citep{madureira-schlangen-2023-instruction}. 
These policies are failing to fully capture the human behaviour from data, but the reasons as still obscure.  

Another open issue is how to collect high-quality CR data in enough amounts for machine learning purposes. In the annotated Minecraft Dialogue Corpus \citep{narayan-chen-etal-2019-collaborative,shi-etal-2022-learning}, TEACh dataset \citep{padmakumar2022teach,gella-etal-2022-dialog} and CoDraw \citep{kim-etal-2019-codraw,madureira-schlangen-2023-instruction,madureira2023are}, CRs occur by own initiative of the players in real, multi-turn interaction, ranging from hundreds to less than ten thousand identified CR utterances. Still in the same size range, the IGLU dataset \citep{kiseleva2022iglu,iglu2022data} has been collected in a setting that avoids pairing up players, with a one-shot opportunity to ask for clarification (and without a partner to answer it and allow further actions). 

Other procedures have been used to collect CR data in larger amounts. Massive datasets are DialFRED \citep{gao2022dialfred}, created via crowdsourcing with workers who are explicitly asked to generate a question, and answer it, for a situation they are not actually involved with. In neighbour domains like virtual assistance, conversational search and code generation, large-scale datasets containing CRs have been constructed with data augmentation methods \citep{aliannejadi-etal-2021-building}, user simulation \citep{kottur-etal-2021-simmc}, templates \citep{li-etal-2023-python} and crawling QA online forums \cite{rao-daume-iii-2018-learning,kumar-black-2020-clarq}. These strategies can reflect CR form and facilitate data collection but abstract away the fundamental triggers of Instruction CRs (joint effort, real-time interaction and action-taking), being arguably not suitable for learning CR policies for instruction following.

\paragraph{Evaluating CR mechanisms in dialogue models} 

We need more evaluation campaigns and methods to shed light on what a model has actually learnt with respect to CR strategies and why it fails. Some initiatives towards more detailed assessment are in progress. \citet{chiyah-garcia-2023-crs} evaluate the abilities of multimodal models to process CRs in coreference resolution by interpreting the difference in the object-F1 score at turns before and after a CR as the improvement provided by incorporating the clarification; they also analyse results by considering various CR properties. In the realm of LLMs, recent studies have employed evaluation techniques via prompts to test the models abilities, concluding that they can detect ambiguity to some extent but even so do not generally attempt to repair it and when they do request clarification there is little alignment with human strategies \citep{kuhn2022clam,shaikh2023grounding}. When \citet{deng2023prompting} first induce the LLM to predict whether the appropriate dialogue act is to ask for clarification
the best LLM achieves only 0.28 F1 Score.

\section{Definitions}
\label{sec:definitions}
CoDraw \citep{kim-etal-2019-codraw} is a multimodal dialogue game where an instruction follower (IF) uses a gallery of 28 (out of 58) cliparts to reconstruct a scene (from the Abstract Scenes dataset \citep{zitnick2013bringing}) they cannot see. They exchange text messages in a turn-based fashion with an instruction giver (IG), who sees the original scene but has no access to the state of the reconstructed scene, except for one chance to peek at it during the game. The available actions are adding or deleting, moving, flipping and resizing cliparts in a canvas. Game success is measured by a scene similarity score based on its symbolic representation. The authors collected 9.9k such dialogues in English, containing around 8k iCRs (11.3\% of the game turns), annotated by \citet{madureira-schlangen-2023-instruction,madureira2023are} both under the license \href{https://creativecommons.org/licenses/by-nc/4.0/}{CC BY-NC 4.0}.

Note that not all iCRs are \textit{questions}. In terms of mood, most CoDraw-iCRs are polar questions, followed by wh- and alternative questions, but there are also declarative and imperative forms. Almost 60\% of instances refer to only one object and around 33\% refer to two objects. The attributes being clarified are, in order of frequency, relations between objects, positions in the scene, disambiguation of persons, direction, size and disambiguation ob objects \citep{madureira2023are}.

We can split the space of possible IF models for this game regarding their CR capabilities:

\vspace{0.3cm}
\noindent \textbf{1. Overhearer}: \textit{A model that observes the current game state (dialogue context and scene) to predict when to ask iCRs, without any additional game-play actions or linguistic decisions.}

\noindent \textbf{2. Action-Taker}: \textit{A model that plays the game by only taking clipart actions, without iCR decisions.}

\noindent \textbf{3. iCR-Action-Taker}: \textit{An Action-Taker with the extra decision of when to ask iCRs.}

\noindent \textbf{4. Full agent}: \textit{A model that makes all game-play decisions, including natural language generation.}
\vspace{0.3cm}

The Overhearer is a common paradigm in NLP in which models resemble an observer of the actual player, deciding what to do \textit{as if it were in their shoes}. It is, however, a rather rough simplification of a full-fledged agent, which is an idealised target not yet reached. (iCR-)Action-Takers are an intermediate step examined in this work.

\paragraph{Task 1} We follow the formalisation of the task of \textit{when to ask} for iCRs in CoDraw by \citet{madureira-schlangen-2023-instruction}. In short, given the game state up to the last IG utterance, the IF has to decide whether to ask for clarification. This policy is modelled as a function $f_{when}: s \mapsto [0, 1]$ that maps the game state $s_t$ at the current turn $t$ to the probability of asking an iCR at this point, performing a binary decision task at each turn in the game. Here, the state $s$ comprises the dialogue history, the gallery and the situation of the scene.

\paragraph{Task 2} Additionally, once the decision to ask has been made, a player should also know what objects are subject to clarification at that point. We thus define the subsequent task of \textit{what to ask} about: at an iCR turn $t$, a function $f_{what}: (o_i, s) \mapsto [0, 1]$ outputs, for each of the 28 objects $o_i$ in the gallery, the probability of asking an iCR about it, given the state $s_t$. These are binary decisions over each available object in the gallery. Both of these tasks are steps happening before the actual generation, which we do not address in this work.\footnote{We leave the additional decisions of what \textit{attributes} to mention and which \textit{form} to realise for ongoing parallel work dealing specifically with iCR generation.}

\section{Hypotheses}
\label{sec:hypotheses}
In this section, we motivate and state the three hypotheses we test as our main contribution. We refer to related findings in the Minecraft game, but note that CoDraw has a more challenging asymmetry regarding the players' common ground: the IG does not observe the IF's actions throughout the game.

\citet{chiyah-garcia-2023-crs} argue that auxiliary learning objectives of detecting objects' attributes in a scene \citep{lee-etal-2022-learning} are useful for referential CRs at Clark's 3rd level, elicited during reference resolution.\footnote{CoDraw-iCR also contains referential CRs, but directly related to uptake of instructions.}
Our expectation is that action prediction should be equivalently relevant for 4th level iCRs, which emerge when deciding how to act. More concretely, iCR-Action-Takers should have a more genuine motivation to decide to request clarification in comparison to Overhearers.\footnote{Experiments in the Minecraft dataset point to the opposite direction: Generating action sequences slightly harmed the accuracy on \textit{when to ask} \citep{shi-etal-2022-learning}. We seek to dive deeper into understanding this issue.} To investigate it, our first hypothesis is:

\begin{hyp} \label{hyp:1}
iCR-Action-Takers can learn a more accurate policy for predicting \textbf{when to ask} an iCR than Overhearers.
\end{hyp}

Here, we can also test whether action \textit{detection} has a similar effect, by letting the model learn to detect actions given the scene before and after, as in \citet{rojowiec-etal-2020-generating}. It is a framing even more equivalent to \citet{lee-etal-2022-learning}, since, in their model, the attributes are already available in the images. The access to post-action scene can be examined in this dialogue game because it is turn-based: The IF would have done all actions they want (thus seeing the newly edited scene) at the point they press the button to send the next message or iCR.

Next, we aim to investigate if Action-Takers, which are trained without any explicit iCR signal, still build representations that encode the need for repair. The study done by \citet{xiao2019quantifying} on quantifying uncertainty in NLP tasks shows that the examined models output higher data uncertainties for more difficult predictions. Besides, \citet{yao-etal-2019-model} propose the assumption that if a model is uncertain about a prediction, it is more likely to be an error, and use uncertainty as a score to decide whether the prediction requires user clarification in semantic parsing. Based on that, we conjecture that the need for repair should manifest as less certainty in the Action-Taker's decisions. Therefore, the second hypothesis we test is:

\begin{hyp} \label{hyp:2}
At iCR turns, Action-Takers predict actions with less certainty than at other turns. Similarly, less certainty is expected for actions upon objects subject to iCRs than for other objects.
\end{hyp}

For this step, we set the linking hypothesis that certainty is expressed in the probability the model assigns to taking action, or not, at a given turn. It is a reasonable assumption, because the objective function is expected to push the predictions to be either 0 or 1, so predictions close to 0.5 can be seen as indecisive.\footnote{An investigation of the predictive uncertainty of the IF model in the Minecraft data has been done by \citet{van2022learning} using length-normalized log-likelihood and entropy of generated action \textit{sequences}. Negative results are reported in an unpublished short manuscript concluding that uncertainty is not a direct signal for when to ask CRs in their setting.} 

Finally, iCR policies for \textit{when to ask} should be grounded in a fine-grained representation of what exactly is unclear. Thus our last hypothesis is:

\begin{hyp}\label{hyp:3}
Pre-trained iCR-Action-Takers can learn a more accurate policy for predicting \textbf{what to ask about} in iCR turns than Overhearers.
\end{hyp}

\section{Models}
\label{sec:models}
In this section, we present the models we analyse in our experiments. We do not intend to propose a novel architecture, since our aim is to understand why current SotA models are failing and the effect that learning to take actions has on them. We implement a model that addresses the limitations of the baseline model (iCR-baseline) from \citet{madureira-schlangen-2023-instruction} by incorporating techniques from top-flight models in recent multimodal dialogue challenges, namely IGLU \citep{kiseleva2022iglu} and SIMMC 2.0 \citep{kottur-etal-2021-simmc}. The basic architecture of the Overhearer and (iCR-)Action-Taker is illustrated in Figure \ref{fig:model}. We provide here an overview of its informationv flow; see Appendix for detailed specifications.

The CoDraw IF has access to a gallery of 28 objects, which is an informative source in the game (\textit{e.g.}~if it contains just one of the three tree cliparts, it is less likely that disambiguation is needed) but was absent in iCR-baseline. We follow a symbolic approach to represent the objects' attributes (presence in the scene, orientation, position, size, pose, facial expression) based on the original drawer in \citet{kim-etal-2019-codraw} (which, however, had unrealistic access to all possible objects in the database).

Previous works did not employ Transformers \citep{vaswani2017attention} to model iCR policies in CoDraw. Given its leading performance in several scenarios, we bring them to the scene, in an approach inspired by DETR \citep{carion2020end}. We use a Transformer decoder\footnote{The full Transformer encoder-decoder was detrimental in almost all cases, so we report results using only the decoder component. This is probably due to the fact that the scene and dialogue had already been encoded by the pretrained components.} module to create contextual embeddings of each object in the current game state, \textit{i.e.}~by building a representation that considers the dialogue so far and the actual scene. 

This is done by passing each object to the Transformer decoder (``target''), to allow self-attention to the state of the gallery, and subsequent cross-attention with the game state representation (``memory''). The state has two components: The dialogue so far, represented via token-level contextual embeddings constructed by BERT \citep{devlin-etal-2019-bert}, and the current scene, represented as image features constructed by a ResNet \citep{resnet50} backbone, followed by a trainable convolutional layer to reduce the number of channels, as in the DETR model \citep{carion2020end}. We make text and scene available as one sequence like \citet{lee-etal-2022-learning}. The variation of iCR-Action-Detecters access the scene before and after the actions.

\begin{figure}[t!]
    \centering
    \frame{\includegraphics[trim={0 6cm 13.5cm 0.2cm},clip,width=\linewidth]{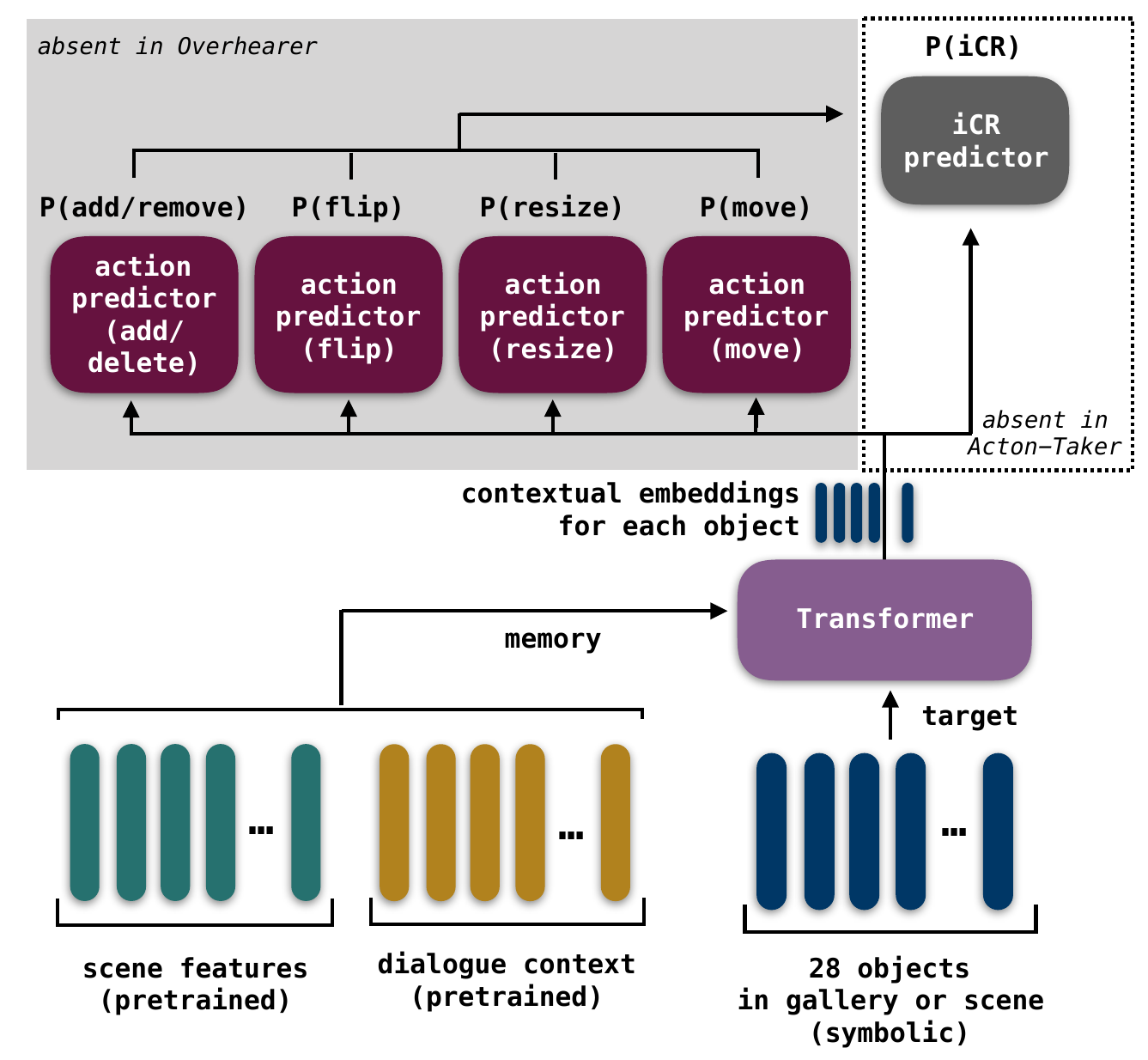}}
    \caption{The basic structure of our iCR policy models. The full structure represents the iCR-Action-Taker. The Overhearer contains no action predictor (area shaded in grey), whereas the Action-Taker contains no iCR predictor (area in the dotted box).}
    \label{fig:model}
\end{figure}

The Transformer outputs a contextual representation of each object. The steps so far are represented in the lower portion of Figure \ref{fig:model}. Now we proceed to the predictions in the upper part, which differs according to the type of model. To test our hypotheses, we implement models that predict the game actions (or detect them, if the updated image is used) and/or make iCR decisions via multi-task learning. We take inspiration from \citet{shi-etal-2022-learning} to train the contextual object embeddings as joint encodings for all the classifiers. 

Action predictors and iCR predictors are implemented as 2-layer feed-forward networks with dropout, which take a representation as input and output a probability. In (iCR-)Action-Takers, we model each action prediction (add/delete, flip, resize, move) as a binary classification done upon each object embedding.\footnote{To facilitate evaluation, we add an additional meta-action prediction which is 1 whenever \textit{any} action is made to a clipart.} 
The iCR decision is also performed as a binary classification task. In Task 1 (when to ask), it predicts whether an iCR should be made at the current turn. In Task 2, (what to ask) it predicts, for each object, whether it is subject to and iCR. In iCR-Action-Takers, we let the action logits be part of the input to the iCR predictor.

\section{Experiments}
\label{sec:experiments}
For our experiments, we implement variations of Overhearers and (iCR-)Action-Takers, all trained on the CoDraw dataset. Results are compared by varying the complexity of the input, which can be comprised of the gallery $G$, the dialogue context $D$ with varying length, the scene before $S_b$ and after $S_a$ the current actions, and the actual actions $A$ or their logits $L_A$.

To test H\ref{hyp:1}, we compare Overhearers with iCR-Action-Takers and iCR-Action-Detecters in Task 1, predicting \textit{when to ask} iCRs at turn level. For H\ref{hyp:2}, we examine the predictions of the Action-Taker using the certainty measure we discuss next. Finally, H\ref{hyp:3} is tested by a similar analysis as H\ref{hyp:1}, but in Task 2, \textit{i.e.}~\textit{what to ask} about. Here, iCR predictions are done at clipart level and only the turns where iCRs actually occurred are used (\textit{i.e.}, we assume the decision to ask for iCR has already been taken). For H\ref{hyp:3}, Overhearers are compared with pretrained iCR-Action-Takers/Detecters whose action modules' parameters are initialised with the best Action-Taker/Detecter checkpoint.

\begin{table}[ht!]
    \centering
    \small
    {\setlength{\tabcolsep}{2pt}
    \begin{tabular}{r%
                    >{\centering\arraybackslash}p{0.8cm}%
                    >{\centering\arraybackslash}p{0.8cm}%
                    >{\centering\arraybackslash}p{0.8cm}%
                    >{\centering\arraybackslash}p{0.8cm}%
                    >{\centering\arraybackslash}p{0.8cm}%
                    >{\centering\arraybackslash}p{0.8cm}%
                    >{\centering\arraybackslash}p{0.8cm}}
        \toprule
         & \multicolumn{2}{c}{\textbf{iCRs}} & \multicolumn{5}{c}{\textbf{actions}}    \\
         \cmidrule(r){2-3} \cmidrule(l){4-8}
         &  when &  what &  any  & add/del & move  & flip  &  resize\\
        \cmidrule(r){2-3} \cmidrule(l){4-8}
             \textbf{train} &  11.24 &   14.32    &  5.43     & 3.11      & 2.13     & 0.23     &0.42     \\
                 \textbf{val} &  11.84 &  14.43     &  5.47     & 3.11      & 2.17     & 0.24     &0.39     \\
                       \textbf{test} &  11.26     &   14.69    &  5.40     & 3.12      & 2.11     & 0.21     &0.39     \\
        \bottomrule
    \end{tabular}
    }
    \caption{\% of the positive labels in the dataset.}
    \label{tab:data-stats}
\end{table}
\begin{table*}[ht!]
    \centering
    \footnotesize
    {\setlength{\tabcolsep}{3pt}
    \begin{tabular}{r l cccccc c cccccc}
    \toprule
      &   & \multicolumn{6}{c}{\textbf{Task 1: When to Ask}} & & \multicolumn{6}{c}{\textbf{Task 2: What to Ask}}  \\
    \cmidrule(lr){3-8} \cmidrule(lr){10-15}
     &  \hspace{1cm} predictions: & \multicolumn{3}{c}{iCR} & \multicolumn{3}{c}{actions} & & \multicolumn{3}{c}{iCR} & \multicolumn{3}{c}{actions}  \\
    \cmidrule(lr){3-5} \cmidrule(lr){6-8} \cmidrule(lr){10-12} \cmidrule(lr){13-15}
     &   inputs    & AP & bF1 & mF1 & AP & bF1 & mF1 & & AP & bF1 & mF1 & AP & bF1 & mF1 \\
    \cmidrule(r){1-2} \cmidrule(lr){3-5} \cmidrule(lr){6-8} \cmidrule(lr){10-12} \cmidrule(lr){13-15}

\textbf{Baseline} & D, $S_a$ &  .347   &  -  &   .645 &            - &             - &             - &     &    - &    - &   - &            - &             - &             - \\

\cmidrule(r){2-2} \cmidrule(lr){3-5} \cmidrule(lr){6-8} \cmidrule(lr){10-12} \cmidrule(lr){13-15}
\textbf{Overhearer} & G &    .138 &       .000 &     .470 &            - &             - &             - &     &    .332 &     .289 &     .593 &            - &             - &             - \\
                    & G, D &    .384 &     .349 &     .642 &            - &             - &             - &     &    .697 &     .665 &     .801 &            - &             - &             - \\
                    & G, D, $S_b$ &    .372 &     .267 &     .604 &            - &             - &             - &     &    .697 &     .666 &     .799 &            - &             - &             - \\
                    & G, D, $S_b$, $S_a$ &    .378 &     .304 &     .620 &            - &             - &             - &     &    .694 &     .660 &     .799 &            - &             - &             - \\
                    & G, D, A &    .372 &     .404 &     .662 &            - &             - &             - &     &    .711 &     .683 &     .810 &            - &             - &             - \\
                    & G, D, $S_b$, A &    .379 &     .377 &     .654 &            - &             - &             - &     &    .712 &     .675 &     .808 &            - &             - &             - \\
                    & G, D, $S_b$, $S_a$, A &    .388 &     .377 &     .655 &            - &             - &             - &     &    .706 &     .674 &     .808 &            - &             - &             - \\

\cmidrule(r){2-2} \cmidrule(lr){3-5} \cmidrule(lr){6-8} \cmidrule(lr){10-12} \cmidrule(lr){13-15}
\textbf{Action-Taker} & G &        - &         - &         - &        .149 &         .005 &         .498 &     &        - &         - &         - &            - &             - &             - \\
                    & G, D &        - &         - &         - &        .769 &         .710 &         .853 &     &        - &         - &         - &        .571 &         .550 &         .770 \\
                    & G, D, $S_b$ &        - &         - &         - &        .762 &         .708 &         .851 &     &        - &         - &         - &        .547 &         .530 &         .761 \\

\cmidrule(r){2-2} \cmidrule(lr){3-5} \cmidrule(lr){6-8} \cmidrule(lr){10-12} \cmidrule(lr){13-15}
\textbf{iCR-Action-Taker} & G, D &    .378 &     .393 &     .658 &        .755 &         .702 &         .848 &     &   \cellcolor[gray]{.9} \textbf{.753} &   \cellcolor[gray]{.9}  .688 &   \cellcolor[gray]{.9}  \textbf{.815} &    \cellcolor[gray]{.9}    .652 &     \cellcolor[gray]{.9}    .621 &    \cellcolor[gray]{.9}     .807 \\
                    & G, D, $L_A$ &    .393 &     .372 &     .652 &        .764 &         .708 &         .851 &     &   \cellcolor[gray]{.9} .751 &   \cellcolor[gray]{.9}  .683 &  \cellcolor[gray]{.9}   .811 &     \cellcolor[gray]{.9}   .657 &     \cellcolor[gray]{.9}    .619 &     \cellcolor[gray]{.9}    .806 \\
                    & G, D, $S_b$ &    .384 &     .380 &     .655 &        .760 &         .702 &         .848 &     &  \cellcolor[gray]{.9}  .739 &  \cellcolor[gray]{.9}   .681 &  \cellcolor[gray]{.9}   .810 &    \cellcolor[gray]{.9}    .612 &    \cellcolor[gray]{.9}     .592 &     \cellcolor[gray]{.9}    .792 \\
                    & G, D, $S_b$, $L_A$ &    .378 &     .311 &     .625 &        .771 &         .709 &         .852 &     &  \cellcolor[gray]{.9}  .743 &  \cellcolor[gray]{.9}   .684 &   \cellcolor[gray]{.9}  .812 &    \cellcolor[gray]{.9}    .630 &    \cellcolor[gray]{.9}     .600 &    \cellcolor[gray]{.9}     .796 \\

\cmidrule(r){2-2} \cmidrule(lr){3-5} \cmidrule(lr){6-8} \cmidrule(lr){10-12} \cmidrule(lr){13-15}
\textbf{iCR-Action-Detecter} & G, D, $S_b$, $S_a$ &    \textbf{.416} &     \textbf{.418} &     \textbf{.676} &        .859 &         .763 &         .880 &     &  \cellcolor[gray]{.9}  .733 &    \cellcolor[gray]{.9} .684 &   \cellcolor[gray]{.9}  .811 &    \cellcolor[gray]{.9}    .834 &     \cellcolor[gray]{.9}    .730 &     \cellcolor[gray]{.9}    .862 \\
                    & G, D, $S_b$, $S_a$, $L_A$ &    .409 &     .366 &     .652 &        .864 &         .777 &         .886 &     &  \cellcolor[gray]{.9}  .739 &   \cellcolor[gray]{.9}  \textbf{.689} &   \cellcolor[gray]{.9}  .813 &    \cellcolor[gray]{.9}    .838 &    \cellcolor[gray]{.9}     .738 &     \cellcolor[gray]{.9}    .867 \\

    \bottomrule
    \end{tabular}
    }
    \caption{Main results of average precision, binary F1 Score and macro-average F1 Score for all models in the test set. The inputs are $G$: gallery, $D$: dialogue, $S_b$: scene before the actions, $S_a$: scene after the actions, $A$: last gold actions, $L_A$: predicted logits of the actions. Shaded cells means the models were pre-trained on actions.}
    \label{tab:main-results}
\end{table*}

Table \ref{tab:data-stats} shows the proportion of each type of label in the dataset. Actions at each turn are sparse (mean=1.65, std=1.69) because only a small subset of the full action space is actually performed.

\paragraph{Implementation} Our implementation uses PyTorch Lightning. We run hyperparameter search and other manual combinations, and then use the configuration that led to the best results in the validation set for the Overhearer G+D model. The training objective is to minimise a sum of binary cross-entropy losses for each task. Optimisation relies on the Adam algorithm \citep{kingma2014adam}, with early stopping. 
Details of the model configuration, data processing and experiment setup are in the Appendix. Our code is available at \url{https://github.com/briemadu/icr-actions}.

\paragraph{Evaluation metrics} We report test results for the best epoch in the validation set.\footnote{We compared Overhearers using a context from 0 to 5 previous turns. 0 or 1 turns had worse results, but 2 to 5 were almost equivalent, so we report results using 3.} H\ref{hyp:1} and H\ref{hyp:3} are analysed based on the performance on iCR predictions. To facilitate comparison to existing works, we report Average Precision (AP) and binary and macro-average F1-Score (bF1 and mF1) for each model and task (\textit{i.e.} one measure for iCR labels and one for all action labels). 
To inspect how much information can be extracted from clipart states alone (\textit{e.g.}~some cliparts are less often subject to iCRs), we report metrics for a model that only gets the gallery as input. For H\ref{hyp:2}, we need an additional prediction certainty metric. We adapt the classification margin metric used for uncertainty sampling in active learning \citep{active-settles}, which is the difference between the probability assigned to the first and the second class, like in \citet{chi2020just}. In our binary task, we define it as $|P(iCR) - P(\neg iCR)|$, which is 0 when both are 0.5 (highest uncertainty) and 1 when one or the other is 1 (highest certainty). We analyse whether we can derive a signal for \textit{when to ask} iCRs by finding a decision threshold upon this metric, as in similar works \citep{yao-etal-2019-model,van2022learning,khalid2023investigating}. 

\section{Results}
\label{sec:results}

Table \ref{tab:main-results} presents the main results for all experiments. We begin with overall observations, and then walk through the table to analyse the findings for each hypothesis. In the next section, we discuss the implications of these findings.

Firstly, for deciding \textit{when to ask} an iCR, the base Overhearer achieves 0.38 AP and the highest performance comes from the iCR-Action-Detecter with 0.41. This is noticeably higher than the 0.34 Overhearer baseline in \citet{madureira-schlangen-2023-instruction}, but the gain is not as substantial as expected given the improvements in the architecture.\footnote{Note that we use the second released version of the annotation, containing a marginally different proportion of iCRs.} When the Overhearer is ablated to have no access to the dialogue, performance drops to close to random, as expected. The addition of scenes before and after the current actions and the inclusion of an explicit signal with the last actions, however, cause only marginal variation and do not really contribute to a better performance. The Action-Taker similarly does not profit from having access to the image. We have no precedent results for the task of \textit{what to ask} about, but even the Overhearer achieves more than .70 AP. Given the imbalance of the labels, we consider it a favourable result, showing this task is easier to model. Introducing iCR decisions does not cause drastic changes to the performance on taking actions for \textit{when to ask}, but fine-tuning on \textit{what to ask} causes a drop, which is probably due to the fine-tuning occurring only on iCR turns. See Appendix for additional analysis.

\paragraph{Hypothesis \ref{hyp:1}} In H\ref{hyp:1}, we study the effect of action-taking on the decision of \textit{when to ask} iCRs. To analyse it, we compare the results of the Overhearer with the iCR-Action-Taker/-Detecter in the left block of Table \ref{tab:main-results}. Integrating multi-task learning for taking actions is slightly helpful for iCR prediction only if the action decision logits are passed to the iCR classifier. If instead of \textit{predicting} actions we let the model learn the auxiliary task of just \textit{detecting} them from the scenes, the results are better.\footnote{Again, this is still plausible: In CoDraw, we can assume that the actual player has taken actions before generating the iCR, as discussed by \citet{madureira-schlangen-2023-instruction}.} Interestingly, the magnitude of the positive difference is comparable to the difference (in accuracy) found in the Minecraft dataset \citep{shi-etal-2022-learning}, which was, however, negative. These effects are not large enough to provide us with definite evidence that H\ref{hyp:1} holds. 

\paragraph{Hypothesis \ref{hyp:2}} For H\ref{hyp:2}, we examine the certainty scores assigned by the Action-Taker to performing \textit{any} action upon each clipart. For the task of \textit{what to ask} about, we compare two distributions: Scores of cliparts subject to iCRs \textit{versus} scores of cliparts not subject to iCRs. For \textit{when to ask} iCRs, we inspect the distributions of the lowest score at turns where iCRs occur \textit{versus} turns where no iCR is made. Using the two-sample Kolmogorov-Smirnov test \citep{hodges1958significance}, we compare the underlying empirical cumulative distributions of the two samples, shown in Figure \ref{fig:example-icr}, under the null hypothesis that they are equal, and a two-sided alternative.

\begin{table}[ht!]
    \centering
    \footnotesize
    {\setlength{\tabcolsep}{2.5pt}
    \begin{tabular}{r cc cc}
    \toprule
     & \multicolumn{2}{c}{\textbf{clipart (what to ask)}} &  \multicolumn{2}{c}{\textbf{turn (when to ask)}} \\
     \cmidrule(lr){2-3} \cmidrule(lr){4-5} 
    & \textbf{iCR} & \textbf{non-iCR} & \textbf{iCR} & \textbf{non-iCR} \\
    \cmidrule(lr){2-3} \cmidrule(lr){4-5} 
    mean (std)     & .838 (.251) & .952 (.147)   & .363 (.283)  & .525 (.328)      \\

    \cmidrule(lr){2-3} \cmidrule(lr){4-5}  
    \textbf{KS test}              & \multicolumn{2}{c}{.524* }        & \multicolumn{2}{c}{.219*}    \\
    \textbf{AP}                      & \multicolumn{2}{c}{.009 \ \ }          &   \multicolumn{2}{c}{.080 \ \ }    \\
    \bottomrule
    \end{tabular}
    }
    \caption{Mean (std) of certainty scores for each sample, results of the two-sided Kolmogorov-Smirnov test and average precision. * means p-value < 0.001.}
    \label{tab:h2-results}
\end{table}

\begin{figure}[!ht]
    \centering
        \includegraphics[trim={0cm 0cm 0cm 6cm},clip,width=0.6\columnwidth]{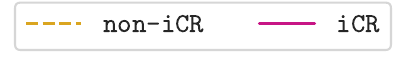}
        \hspace{2cm}
        \includegraphics[trim={0.9cm 1.8cm 1cm 0cm},clip,width=0.45\columnwidth]{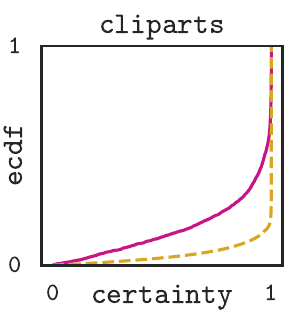}
        \includegraphics[trim={0.9cm 1.8cm 1cm 0cm},clip,width=0.45\columnwidth]{figures/cdf-icr_turn_label-certainty.pdf}

    \caption{Empirical cumulative distribution function of the certainty of taking actions for each clipart (left) and the minimum by turn (right).}
    \label{fig:ecdf}
\end{figure}

Table \ref{tab:h2-results} shows the statistically significant test results. It means that, on the whole, Action-Takers behave differently regarding action certainty for cliparts or turns with iCRs. In Figure \ref{fig:ecdf}, we can see that the certainty for non-iCR cliparts is more concentrated around 1 than for cliparts subject to iCRs. Similarly, the distribution of the minimum certainty score at iCR turns is more concentrated at lower values. In that sense, we find support for H\ref{hyp:2}. Still, using these scores directly as a signal for iCR prediction does not result in high AP, in line with the findings by \citet{van2022learning}. This seems to occur because, although the distributions are different, both samples have values in the whole range, with overlap in their standard deviation. 
 
\paragraph{Hypothesis \ref{hyp:3}} Lastly, we assess the effect of taking actions in deciding \textit{what to ask} about. Here, we focus on the right columns of Table \ref{tab:main-results}, again comparing the Overhearer with the pretrained iCR-Action-Takers/Detecters. We observe a positive effect of learning to take actions on the iCR policy, with AP increasing from .69 to .75. Differently from the task of \textit{when to ask}, here \textit{predicting} actions leads to better results than merely \textit{detecting} them. The difference is not negligible, which is stronger support in favour of H\ref{hyp:3} in this context.

\section{Discussion}
\label{sec:discussion}

Our setting allowed us to differentiate between \textit{understandability} and iCR \textit{policy}. The first refers to learning a mapping from linguistic input to actions. The latter is an additional decision on top of action-taking that regards knowing when the information available to the agent at a given moment is not enough for the current purposes of wanting to commit to an outcome. 

Learning to take actions does not seem to be a signal informative enough for deciding \textit{when to ask} for iCRs, although it has a more prominent effect on deciding \textit{what to ask} about in iCR turns. Besides, we investigated whether there is a signal in the purely understanding models that predicts what to clarify. Indeed, a model trained without any explicit iCR signal made predictions whose certainty distribution differ at iCR turns and cliparts. Even though the raw score cannot be directly used as a predictor of human iCR behaviour, further investigation can be done on extracting an agent's implicit iCR policies, \textit{e.g.}~with probing or attribution methods and in-depth analysis of the model's internal states.

The five sources of improvement (integration of the gallery, token-level representations of utterances, learnable scene features, attention mechanism to construct contextual object embeddings and action predictions) over the existing CoDraw baseline formed together a conceptually superior model design. 
We expected this more sophisticated architecture, aligned with the latest literature, to lead up to a clear-cut improvement in the task of when to ask iCRs. The fact that the gain is not more than 10\% in our main metric over that baseline compel us to join the ranks of works that question whether the current NLP paradigm (employing imitation learning or behavioural cloning to learn with supervision from limited human data) is the right way to go when it comes to meta-discursive acts in interactions \citep[\textit{inter alia}]{hayes-1980-expanding,nguyen2022framework, min-etal-2022-dont, van2022learning, imitation-learning-stanford}. It is also possible that the actions signal is too weak; the action space is large (four actions on 28 objects) which makes the actually performed actions at a given turn be sparse.

In a static dataset of human play, the underlying CR policies of each player may differ by nature and also in visibility in the data. We cannot know with certainty if other humans would have behaved differently at each point than what is realised in the data; consequently, it is hardly possible to set a standard against which to judge the trained model's policy. We are, after all, trying to learn a ``customary'' policy from what is actually a mixture of policies with observations sampled from various players. It may be the case that we have reached the limits of the generalisable policies we can capture from this data with supervised training, even though the actual metrics are not close to the ceiling.\footnote{Though, as pointed out by a reviewer, this may be a limitation of the class of models we tested, and results can possibly be improved with more powerful vision/language encoders.} As \citet{hayes-1980-expanding} discussed, graceful interaction requires developers to aim for non-literal aspects of communication that are effective for the human-agent interaction, instead of trying to imitate human patterns exactly. 
This connects to the over confidence problem in LLMs: In some situations, they should produce an \textit{I don’t know} or a CR, but their limited abilities in meta-semantic communication often cause failures.

Ambiguity arises under competing communicative pressures \citep{piantadosi2012communicative}. Thus CRs are not a problem: They are a solution emerging from joint effort \citep{clark2002speaking}. If many bits of information are to be conveyed, the IG may produce minimally sufficient messages and leave it to the addressee to identify gaps. The IF may also take actions that are only approximately good, since mistakes can normally be fixed later. 
Moreover, crowdworkers seem to lack incentive to try to build perfect reconstructions, and often seem to use implicit knowledge to make only satisfactory actions (see Appendix). Therefore, the iCR signal may not be ``out there'' in the data, but live in the internal state of the agents. Treating the task as \textit{iid} predictions under supervised learning is also not ideal because game decisions are actually made sequentially. Like some works on learning when to ask questions, modelling iCR policies may call for reinforcement learning (see \textit{e.g.}~\citet{khalid-etal-2020-combining}), with evaluation methods that capture the effectiveness of the agent's policy for the game, beyond comparison with human behaviour.

\section{Conclusion}
\label{sec:conclusion}
We have examined the effects of performing actions on learning iCR policies in the CoDraw game. The assumption that learning to take actions would make the underlying \textit{when to ask} policy emerge does not fully hold. Still, we find that prediction certainty of actions differs at iCR turns. Then, if we assume that a given policy has informed us on \textit{when} iCRs have to be made, we show that it is possible to predict \textit{what to ask} about more successfully, with action-taking having a stronger positive effect. Exploring larger datasets with CRs produced as a by-product of action-taking is desired. Still, the suboptimal performance of various SotA models in deciding \textit{when to ask} for clarification speaks against approaches that seek to imitate human behaviour. We recommend more investigation with RL and evaluation methods that capture the effectiveness of iCR policies in dynamic contexts.

\section{Limitations}
\label{sec:limitations}
We have only explored one dataset because there are very few genuine iCR datasets available yet. Minecraft, which is relatively comparable in terms of the underlying instruction following setting, is smaller and has a different form of common ground due to full visibility by the IG. It has been explored in related work, to which we refer in the related work section. SIMMC 2.0 is not suitable in this context for two reasons: Its CRs are not at Clark's level 4 (uptake), but mostly level 3 (reference resolution). Besides, it is a simulated dataset, and we are interested in exploring the limits of modelling human iCR behaviour. 

The models are thus task-specifically fitted to CoDraw and cannot be applied out of the box to other domains. Still, we believe that CoDraw is representative of iCRs and that solving the task in one domain is a first step towards generalisation, which has not been achieved yet even with other datasets, as we discussed.

In this work, our models do not predict all fine-grained game actions, \textit{i.e.}~they are not full-fledged Action-Takers. In preliminary experiments, we first attempted to model an agent that predicts all features of each clipart at each turn. However, since the vast majority of the 28 available cliparts remain unchanged from one turn to the other, the model could simply learn to output a copy of the current state. We thus opted to turn all tasks into binary predictions for our analysis, as we observed results that are good enough for our purposes, given the imbalanced nature of the actions in the data. For each object in the gallery, it makes high level decisions on which actions are needed (add/delete, move, resize, flip). A full agent should include the subsequent tasks of deciding where to place cliparts and what exact (discrete) size to set (presence and orientation can be deduced in post-processing with the current version). 

Further investigation can be done to improve the performance of the Action-Takers. Since the actions are very sparse, it may be the case that models just learn to detect mentioned cliparts in the utterances. A detailed error analysis should look closer at the predictions and also examine how good the scene similarity scores of the reconstructions are. Instead of predicting probabilities, the model could also output parameters of a distribution from which the actions would be sampled; we do not investigate that option here. Besides, we use a supervised learning approach that treats turns as \textit{iid}. In reality, what the player does in one turn influences its next moves, so other methods like RL could be more appropriate, as we discussed. 

Although our models take several epochs to overfit the training data, performance in the validation set saturates very early. The techniques we tried (for instance, dropout, variations of the architecture and filtering the training data) did not lead to better results. We performed a limited hyperparameter search that could be done more extensively in the future, also to investigate in more detail how the method scales with larger and smaller models.

For the task of \textit{what to ask} about, we did not include the utterances for which the annotation does not provide the reference cliparts due to ambiguity. Still, that happens for very few cases and should not have a considerable impact on the results.

To conclude, we do not have human performance to use as an upper boundary for our results. It would be interesting to collect human data by letting humans decide \textit{when to ask} for clarification and \textit{what to ask} about, so that we can better understand to what extent the task itself is possible for humans acting as overhearers. Still, since our aim is to do an intrinsic analysis on whether taking actions improve a model's performance, human results are not strictly necessary, because comparison within models suffices for testing our hypotheses.

\section{Ethical Considerations}
Merely posing clarification requests can be a source of miscommunication regarding intentions, which has ethical implications and may also weaken the application of moral norms by the interlocutors, as discussed by
\citet{jackson2018robot} and \citet{jackson2019language}. Besides, the risks regarding privacy and biases of learning actions from individual behaviour also apply, as well as the current topics being discussed in the field of responsible NLP.

\section*{Acknowledgements}
We thank the anonymous reviewers for their valuable comments that helped improve the paper. We also thank Philipp Sadler and Javier Chiyah-Garcia for helpful discussions regarding this research project.

\bibliography{anthology,custom}

\appendix

\section{Additional Analysis}
\label{sec:appendix-a}
\begin{figure}[!ht]
    \centering
        \includegraphics[trim={0cm 0cm 0cm 0cm},clip,width=0.8\columnwidth]{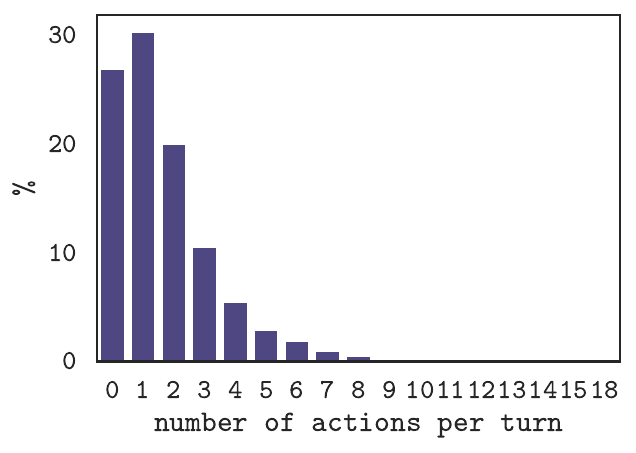}
    \caption{Empirical distribution of the number of actions per turn in the CoDraw dataset.}
    \label{fig:actions-dist}
\end{figure}

\begin{figure}[!ht]
    \centering
        \includegraphics[trim={0cm 0cm 0cm 0cm},clip,width=0.55\columnwidth]{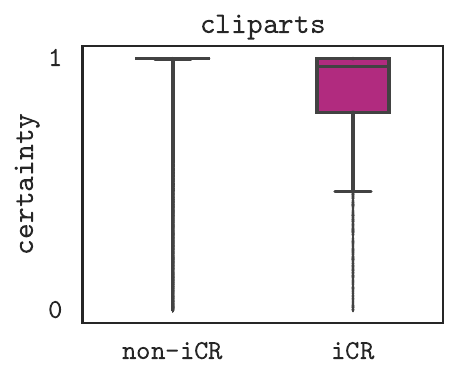}
        \includegraphics[trim={0cm 0cm 0cm 0cm},clip,width=0.55\columnwidth]{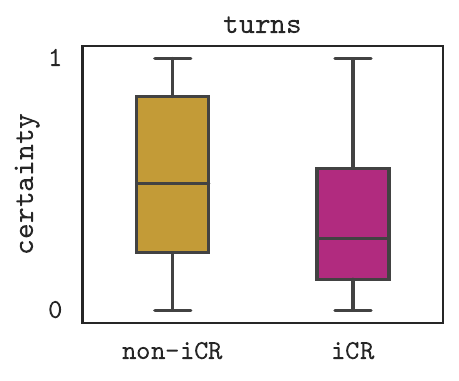}

    \caption{Empirical distribution of the certainty of taking actions for each clipart (top) and the minimum by turn (bottom).}
    \label{fig:boxplots}
\end{figure}

Here we present additional analysis. Figure \ref{fig:actions-dist} illustrates the distribution of the number of actions per turn. Table \ref{tab:actions} presents the average precision for each type of action, which are aggregated in Table \ref{tab:main-results}. Figure \ref{fig:boxplots} show the boxplots for the distribution of certainty scores, to aid visualising that they have different shapes in each sample.

\begin{table}[ht!]
    \centering
    \footnotesize
    {\setlength{\tabcolsep}{1.3pt}
    \begin{tabular}{r l cccc}
    \toprule

    & &  \textbf{add/del} &  \textbf{move} &  \textbf{flip} &  \textbf{resize} \\
    \cmidrule{3-6}
    \textbf{Action-Taker}        & G, D                  &   .875 &  .617 &  .367 &  .531 \\
    \textbf{iCR-Action-Taker}    & G, D                  &   .865 &  .600 &  .398 &  .539 \\
    \textbf{Action-Detecter}     & G, D, $S_{a,b}$ &   .976 &  .644 &  .414 &  .636 \\
    \textbf{iCR-Action-Detecter} & G, D, $S_{a,b}$ &   .974 &  .642 &  .423 &  .626 \\
    \bottomrule
    \end{tabular}
    }
    \caption{Detailed performance of the Action-Takers and Action-Detecters for \textit{when to ask}. Values are the average precision for each type of action in the test set.}
    \label{tab:actions}
\end{table}

\section{Reproducibility}
\label{sec:appendix-b}
In this section, we provide details of our data pre-processing and implementation. For precise details, please check the available code. Here, we provide a brief overview of each component and the justification of some decisions. 

\subsection{Data}

    We used the annotation released in the file \texttt{codraw-icr-v2.tsv}\footnote{\url{https://osf.io/gcjhz/files/osfstorage}} to identify iCRs and mentioned cliparts. We followed the train-val-test splits as in the original CoDraw data. The \textit{ambiguity classes} introduced by the authors were treated as follows: If an iCR was about an ambiguous but concrete class, we assigned the positive iCR label to all objects in the gallery that belong to that class. For instance, for \texttt{hat\_group}, all hats in the gallery were treated as positive cases. The general ambiguity class, used for unclear cases, was ignored in our labelling. This occured in 318 iCRS. The whole dataset was used in all experiments, except for the tasks of \textit{what to ask} about, for which only the turns containing iCRs were included for all splits.

    The gallery and scene representation was constructed using features in a similar fashion as the original paper. Each clipart was assigned integers for its identifier, size (three categories), orientation (two categories), presence in the current scene (a binary feature), pose (seven categories) and facial expression (5 categories), as well as five features for its position (x and y coordinates of its centre, width, height and area in the canvas). We set features (except pose and facial expression) to a special category 0 for objects that are not in the scene. All boy and girl cliparts were collapsed into one class for each, and their facial expressions and poses were turned into features in the symbolic representation, as in original paper. Other cliparts were assigned a ``not-applicable'' class for these two features. To define bounding boxes, we rescaled sizes according to the AbstractScenes documentation.
    
    Actions were defined as either addition/deletion or edits. Edits meant flip, resize and move. If a clipart was added or deleted, we did not consider changes to its orientation, position and size with respect to the gallery (in order to avoid that the model only learnt the edits that occur due to an addition or deletion). Actions were defined by comparing the state of the gallery in a turn in relation to its state in the previous turn. For initial turns and some cases where the scene string was not available in the dataset, we set the scene to empty and use the gallery in adjacent turns (since the gallery should remain the same across the game). We also introduced an ``acted upon'' action that is positive whenever any type of action occurs upon a clipart.
    
    Text embeddings were retrieved from \texttt{bert-base-uncased}, licensed under \href{https://choosealicense.com/licenses/apache-2.0/}{Apache 2.0}. Following \citet{shi-etal-2022-learning}, we concatenate the IG and IF utterances using special tokens before each speaker. Special tokens \texttt{<TELLER>} and \texttt{<DRAWER>} were appended before the instruction giver and follower, respectively. The last utterance from the instruction follower was appended to the beginning of the utterance of the instruction giver, so that potential previous iCRs are encoded with their responses, if given immediately. Embedding sequences were padded with zeros to the right to an empirical length of 80 tokens. When context is used, the previous turns are appended to the left of the last instruction and, if necessary, padded with zeros to the left, so that the most recent turn is always at the same position in the input.

\subsection{Implementation}

    \begin{table*}[ht!]
        \centering
        \small
        \begin{tabular}{p{3.5cm} l l l} 
         \toprule
          \textbf{hyperparameter} & \textbf{type} & \textbf{options} & \textbf{selected} \\ 
         \midrule
            accumulate gradient & discrete & 1, 2, 5, 10, 25 & 1 \\
            batch size & discrete & 16, 32, 64, 128, 256 & 32 \\
            clipping  & discrete & 0, 0.25, 0.5, 1, 2.5, 5 & 1 \\
            context length & integer & min=1, max=5 & 3 \\
            dropout & discrete & 0.1, 0.2, 0.3 & 0.1 \\
            d\_model & discrete & 128, 256, 512 & 256 \\
            hidden\_dim & discrete & 32, 64, 128, 256, 512, 1024 & 256 \\
            hidden\_dim\_trf & discrete & 256, 512, 1024 & 2048 \\
            learning rate & discrete & 0.1, 0.01, 0.001, 0.0001, 0.003, 0.0003, 0.00001, 0.0005 & 0.0001 \\
            lr scheduler & bool & True, False & False \\
            lr step & integer & min=1, max=10 & - \\
            n heads & discrete & 1, 2, 4, 8, 32 & 16 \\
            n layers  & float & min=1, max=6 &  3 \\
            n reload datasets  & float & min=1, max=10 & 1 \\
            pos weight  & float & min=0.8, max=3 & 2 \\
            pre-trained text embeddings & categorical & bert-base-uncased, roberta-base, distilbert-base-uncased & bert-base-uncased \\
            random seed & integer & min=1, max=54321 & 12345 \\
            weight decay & discrete & 1, 0.1, 0.01, 0.001, 0.0001 & 0. \\
            weighted loss & bool & True, False & False \\
        \bottomrule
        \end{tabular}
        \caption{Hyperparameters: Investigated options and selected values. Note that the search did not extensively cover all possibilities for each hyperparameter.}
        \label{tab:hyperparameters}
    \end{table*}

    The models were implemented with Python (v3.10.12), PyTorch\footnote{\url{https://pytorch.org/}} (v1.13.1) and Pytorch Lightning\footnote{\url{https://lightning.ai/pytorch-lightning}} (v2.0.8), in Linux 5.4.0-99-generic with processor x86\_64 on an NVIDIA GeForce GTX 1080 Ti GPU with CUDA (v11.6). The pre-trained ResNet model was retrieved from torchvision\footnote{\url{https://pytorch.org/vision/stable/models.html}} (v0.14.1) and the pre-trained BERT came from HuggingFace transformers\footnote{\url{https://huggingface.co/bert-base-uncased}} (v4.29.2).
    
    Optimisation was done with the Adam algorithm \citep{kingma2014adam}, using \texttt{BCEWithLogitsLoss} with \texttt{reduction} set to sum and the argument \texttt{pos\_weight} to 2 for each task. The total loss used for backpropagation was a sum of all task losses. Early stopping was implemented using a patience of 8 epochs and the minimum delta of 0.001 for maximisation of a monitored metric. Metrics were computed using torchmetrics\footnote{\url{https://torchmetrics.readthedocs.io/en/latest/}}  (v0.11.4). The monitored metric varied according to the task: If iCRs were predicted, we tracked the binary average precision of iCR labels; otherwise, we tracked the binary average precision of the meta-action class. The maximum number of epochs was set to 30. The checkpoint that lead to best performance in the validation set was saved and loaded to run the tests. Comet\footnote{\url{https://www.comet.com}} was used to manage experiments and to perform hyperparameter search.

    Hyperparameter search was performed with the base model (\textit{i.e.}~an Overhearer that gets only the dialogue and the gallery representation as input and predicts only \textit{when to ask} iCRs). We used comet's Bayes algorithm as well as a few manual selections of hyperparameters, and opted for the model with highest iCR binary average precision in the validation set. Table \ref{tab:hyperparameters} shows the final hyperparameter configuration used in all experiments. 
    
    We did not keep records of all experiments during development. For the final run, we run 43 experiments during tuning and 102 for the analysis. The duration varied from 5 minutes (the random baseline) to 06h16m (the iCR-Action-Detecter using the full Transformer), without including the time for data preparation. The number of parameters varied according to the model. The turn-level Overhearer without scenes had 5,008,923 and with both scenes 29,054,299 (5,546,267 learnable). The turn-level iCR-Action-Taker without scenes had 5,339,168, and the iCR-Action-Detecter had 29,384,544 (5,876,512 learnable).
    
    To enable reproducibility, we set the use of use deterministic algorithms to \texttt{True} in PyTorch and used Lightning's \texttt{seed\_everything} method with a fixed random seed. Despite this, according to the documentation, some methods cannot be forced to be deterministic in PyTorch when using CUDA.\footnote{\url{https://pytorch.org/docs/1.13/generated/torch.use_deterministic\_algorithms.html\#torch.use\_deterministic\_algorithms}}

    \subsection{Model}

    In this section, we explain in more detals how we address five of the limitations of the baseline model (iCR-baseline) by \citet{madureira-schlangen-2023-instruction}, some of them already acknowledged by the authors. We also refer to the original CoDraw model (CoDraw-orig) by \citet{kim-etal-2019-codraw}, which, however, did not include the instruction follower's utterances in the game. 
    
    \paragraph{Incorporating the gallery} The gallery is an informative source in CoDraw (\textit{e.g.}~if it contains just one of the three tree cliparts, it is less likely that disambiguation is needed). iCR-baseline does not include the available objects as input, whereas CoDraw-orig uses a symbolic representation assuming all 58 objects are available at any time. Both approaches do not correspond to reality, as players only see 28 cliparts. We follow a similar symbolic approach to represent the objects' attributes (presence in the scene, orientation, position, size, pose, facial expression), but only for those at play. The cliparts' features and bounding boxes are projected to a higher-dimensional space following \citet{sadler-schlangen-2023-pento}.
    
    \paragraph{Using contextual word embeddings} iCR-baseline relies only on two sentence-level embeddings, one to encode the whole dialogue context and one for the last utterance, both not optimised for the game. To allow the policy to access more fine-grained linguistic information, we make all token-level contextual embeddings available to the player, constructed by a pretrained language model. 
    
    \paragraph{Enhancing scene representations} iCR-baseline uses a pretrained image encoder. It is unlikely that off-the-shelf encoders fit well to clipart scenes without fine-tuning. Here, we follow the approach in DETR \citep{carion2020end}, employing a ResNet \citep{resnet50} backbone with learnable positional encodings to extract scene features, followed by a trainable convolutional layer to reduce the number of channels. The sequence of image features is then used as part of the input.
    
    \paragraph{Transforming} The iCR predictions rely only on pretrained embeddings with a feed forward neural network in iCR-baseline, and CoDraw-orig did not employ Transformers \citep{vaswani2017attention} as a trainable component. Given its leading performance in several scenarios, we bring them more explicitly to the scene, in an approach similar to DETR \citep{carion2020end}. We feed the clipart representations to the decoder, to allow self-attention to build up embeddings of the state of the gallery and scene, without positional encoding due to the arbitrary order of the cliparts. Here, we also rely on the findings by \citet{chiyah-garcia-2023-crs} that encoding relations between objects and their locations is helpful for CRs. Then, it performs cross-attention with the scene and text.  
    We make text and scene available as one sequence like \citet{lee-etal-2022-learning}. Since cross-attention between modalities is a cornerstone in current CR models \citep{shi-etal-2022-learning,shi2023and}, we also run experiments using the encoder to let text and scene attend to each other. We then end up with a multimodal representation of each clipart in the current context, which is then passed to classifier layers for each prediction. 
    
    \paragraph{Action-taking via multi-task learning} iCR-baseline is an Overhearer, modelling only the policy of \textit{when to ask} iCRs. To test our hypotheses, we implement (iCR-)Action-Takers that predict the game actions (or detect them, if the updated image is used) via multi-task learning. Note that this is not yet a full-fledged Action-Taker. For each object in the gallery, it makes high level binary classification on which actions are needed (add/delete, move, resize, flip); a full model would also make the subsequent fine-grained decision of exact positions and sizes. We take inspiration from \citet{shi-etal-2022-learning} and train a joint encoding for multiple classifiers. We let the action logits (or the real actions via teacher forcing) be part of the input to the iCR decoder. To facilitate evaluation, we add an additional meta-action prediction which is 1 whenever \textit{any} action is made to a clipart.

    \paragraph{Components} Let d\_model be the dimension used for the Transformer. First of all, an embedding of the gallery and scene state is constructed. Embedding layers are used for a clipart's identifier, orientation, presence, size, face and pose states with dimensions d\_model-100, 10, 10, 10, 20 and 20, respectively. The position is embedded with a linear layer that maps its centre coordinates, area, width and height to 30 dimensions. All embedded features are concatenated so as to create a representation with dimensions 28 (number of cliparts) by d\_model. We used only the decoder of the Transformer, which gets the gallery representation as ``target'' and the instruction tokens (whose dimensions were reduced with a linear layer and, if applicable, the sequence was concatenated to the scene features) summed to positional encodings as ``memory''. The decoder performs self-attention in the gallery and then cross-attention with the memory. Scenes are encoded following \citet{carion2020end}'s implementation, but we first preprocess the scene according to the pretrained model's documentation. The scene is then fed into a pre-trained ResNet50 followed by a trainable convolutional layer that reduces the number of channels to the same dimension used for the Transformer. Then, the height and width dimensions are flattened and the result is added to learnable position embeddings, with a dropout layer. The probabilities (for iCRs or actions) are predicted by taking each output of the Transformer (\textit{i.e.}~one representation for each clipart in the gallery) and passing it through a feed-forward network with the following sequential layers: leaky ReLU, dropout, linear, leaky ReLU and linear. For predicting turn-level iCRs, the representations of all cliparts are averaged. If the action-taking logits or teacher forcing is used, they are appended to the input. The output logits are converted to probabilities using the sigmoid function.

    \subsection{Evaluation}
    
    The threshold for the F1-Scores was set to 0.5. We did not include the meta-action label in the main results for taking actions to avoid inflating the performance; it was only used for the analysis for H\ref{hyp:2}, done on the Action-Taker+G, D. Metrics for the evaluation were computed with sklearn\footnote{\url{https://scikit-learn.org/stable/index.html}} (v1.0.2) and the plots were generated with seaborn (v0.12.2) and matplotlib\footnote{\url{https://matplotlib.org/}} (v3.7.1). The hypothesis test was done with SciPy\footnote{\url{https://docs.scipy.org/doc/scipy/reference/generated/scipy.stats.ks\_2samp.html}} (v1.11.1) \texttt{stats.ks\_2samp} method with a two-sided alternative.

\section{CoDraw Examples}
\label{sec:appendix-c}
\begin{figure*}[!ht]
    \centering
        \includegraphics[trim={1.8cm 6.9cm 5cm 6.5cm},clip,width=\linewidth]{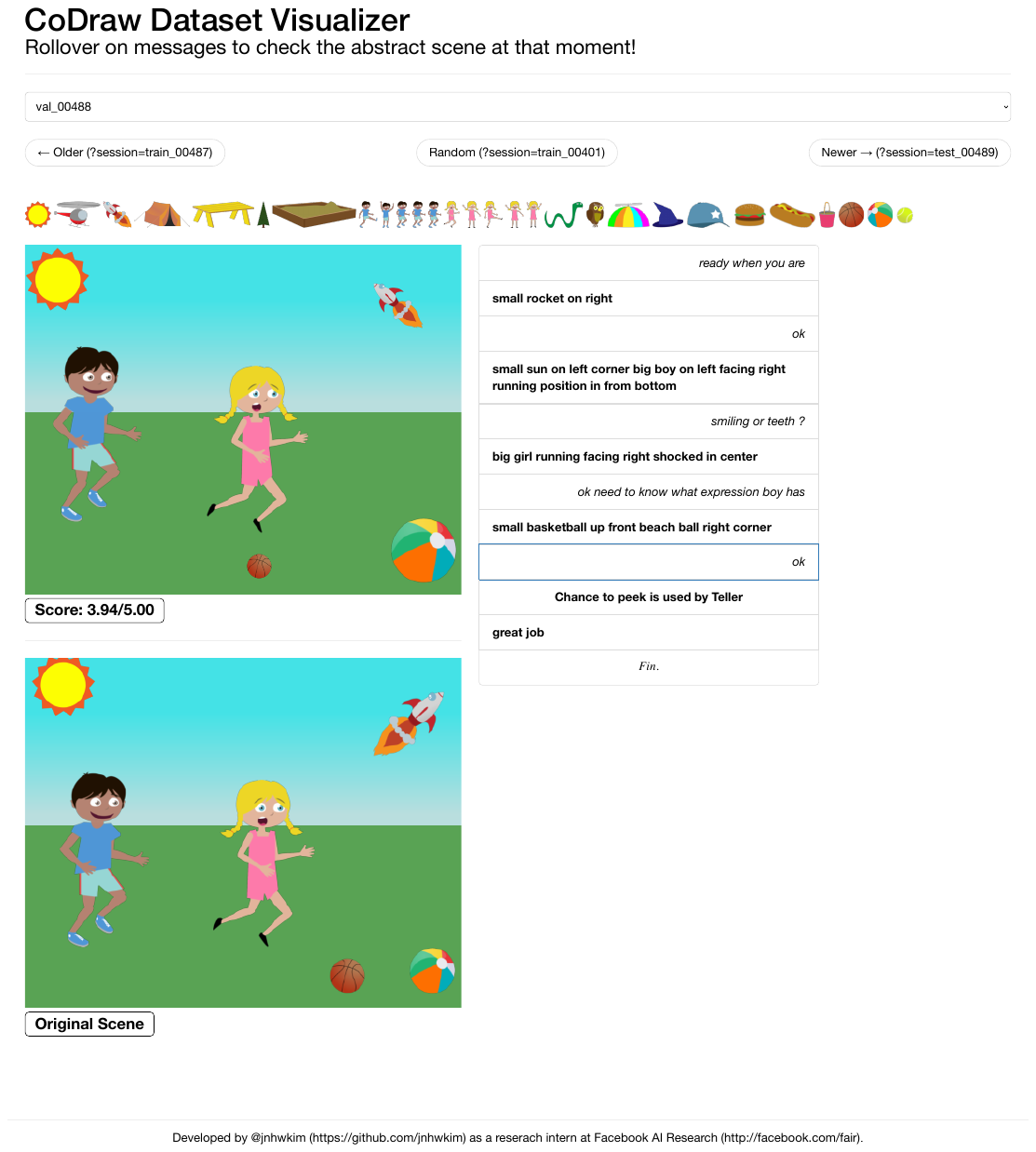}
    \caption{Even peeking, the instruction giver does not inform the instruction follower that the reconstruction is not totally correct: The orientation of the rocket is wrong, as well as the position of the basketball and the size of the two balls. From: CoDraw dialogue game 488, \href{https://creativecommons.org/licenses/by-nc/4.0/}{CC BY-NC 4.0}, scene from \citet{zitnick2013bringing}.}
    \label{fig:codraw-examples-1}
\end{figure*}

\begin{figure*}[!ht]
    \centering
        \includegraphics[trim={1.8cm 6.9cm 5cm 6.5cm},clip,width=\linewidth]{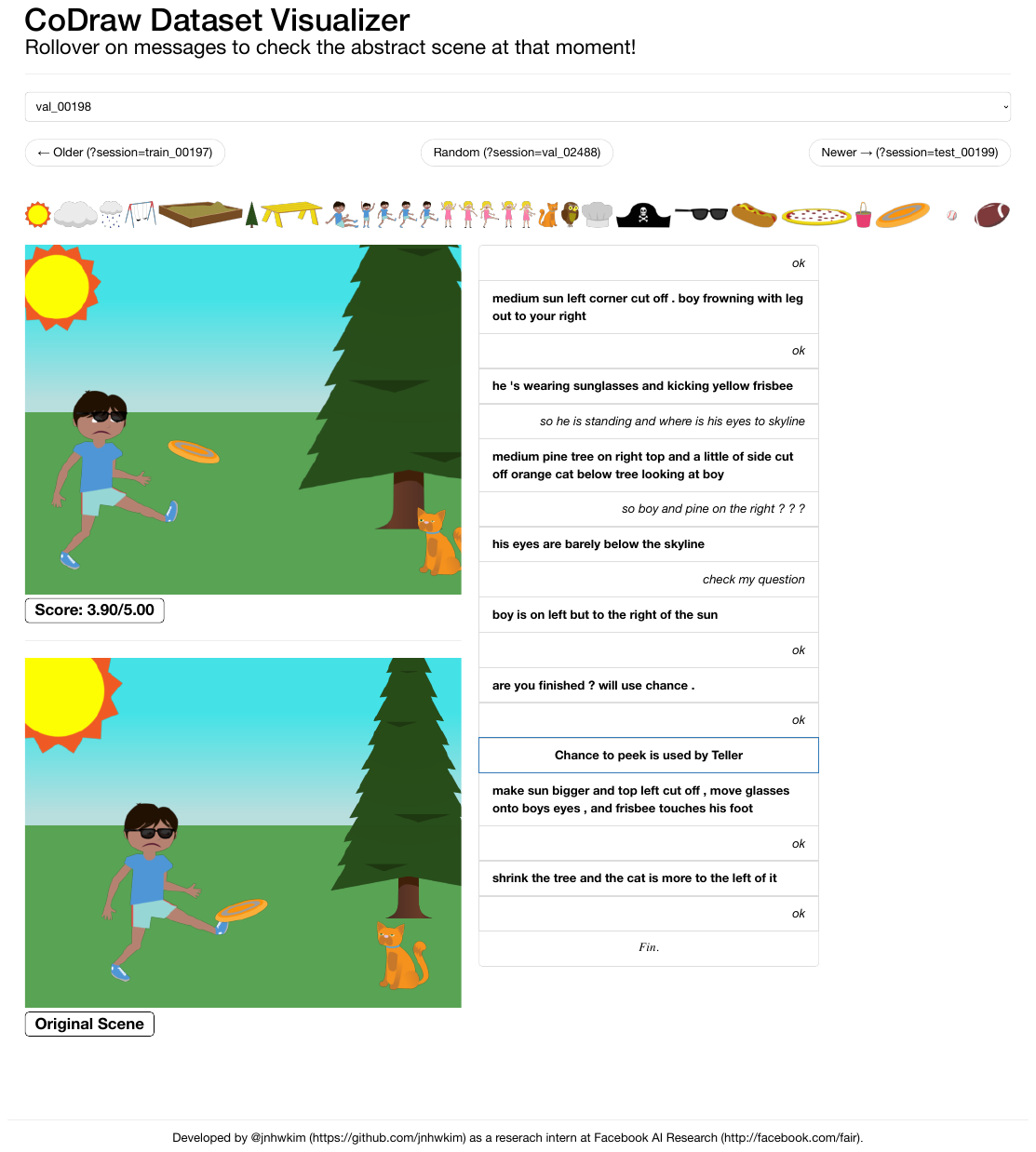}
    \caption{A more careful instruction giver uses two turns to try to repair even minor details after the peek, like the slightly wrong position of the sunglasses. From: CoDraw dialogue game 198, \href{https://creativecommons.org/licenses/by-nc/4.0/}{CC BY-NC 4.0}, scene from \citet{zitnick2013bringing}.}
    \label{fig:codraw-examples-2}
\end{figure*}

\begin{figure*}[!ht]
    \centering
        \includegraphics[trim={1.8cm 6.9cm 5cm 6.5cm},clip,width=\linewidth]{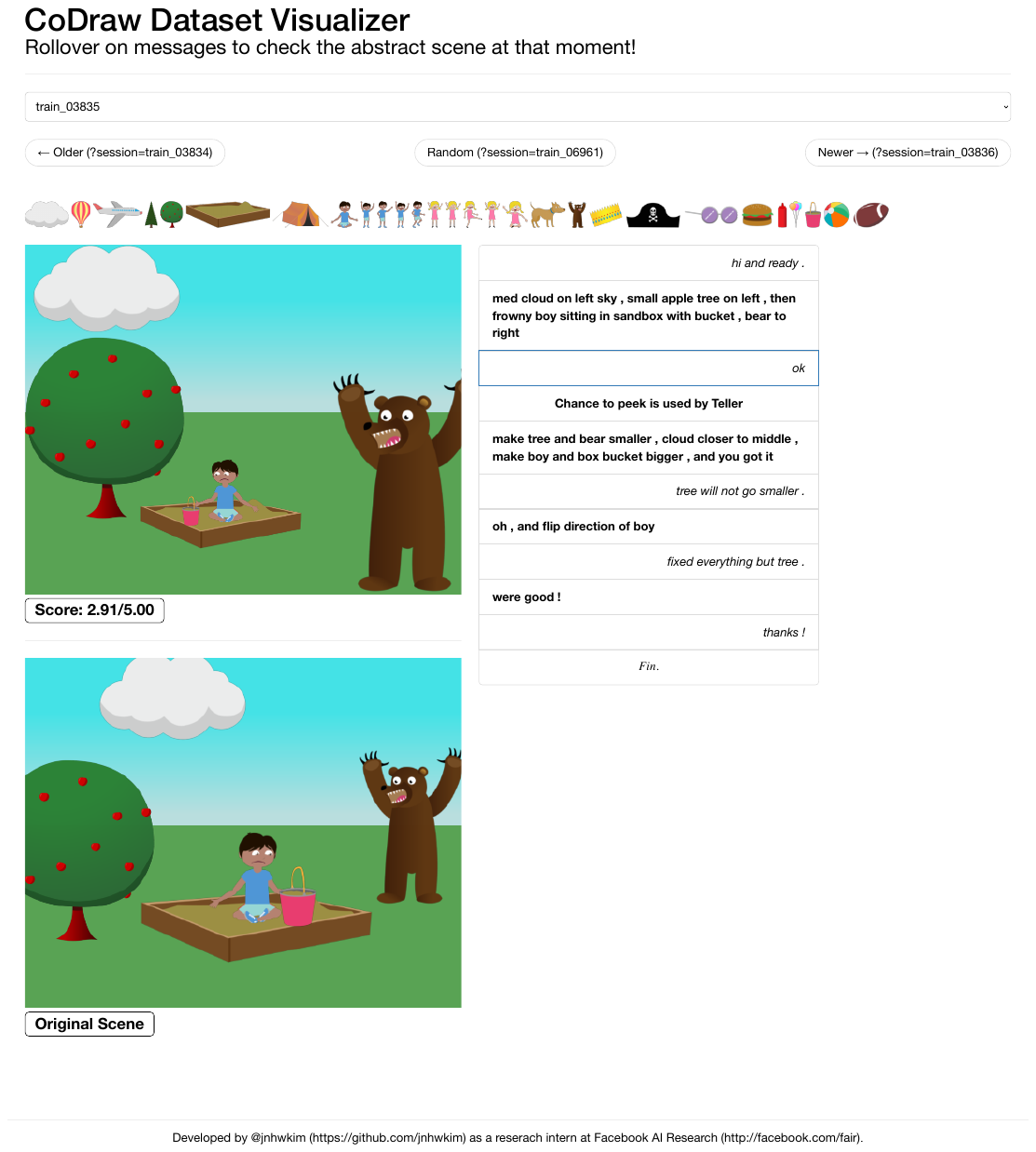}
    \caption{The instruction follower gets underspecified instructions at the first turn (for instance, nothing is said about the orientation of the boy and his position with respect to the bucket), but acts even so without asking for clarification. From: CoDraw dialogue game 3835, \href{https://creativecommons.org/licenses/by-nc/4.0/}{CC BY-NC 4.0}, scene from \citet{zitnick2013bringing}.}
    \label{fig:codraw-examples-3}
\end{figure*}

\begin{figure*}[!ht]
    \centering 
        \includegraphics[trim={1.8cm 6.9cm 5cm 6.5cm},clip,width=\linewidth]{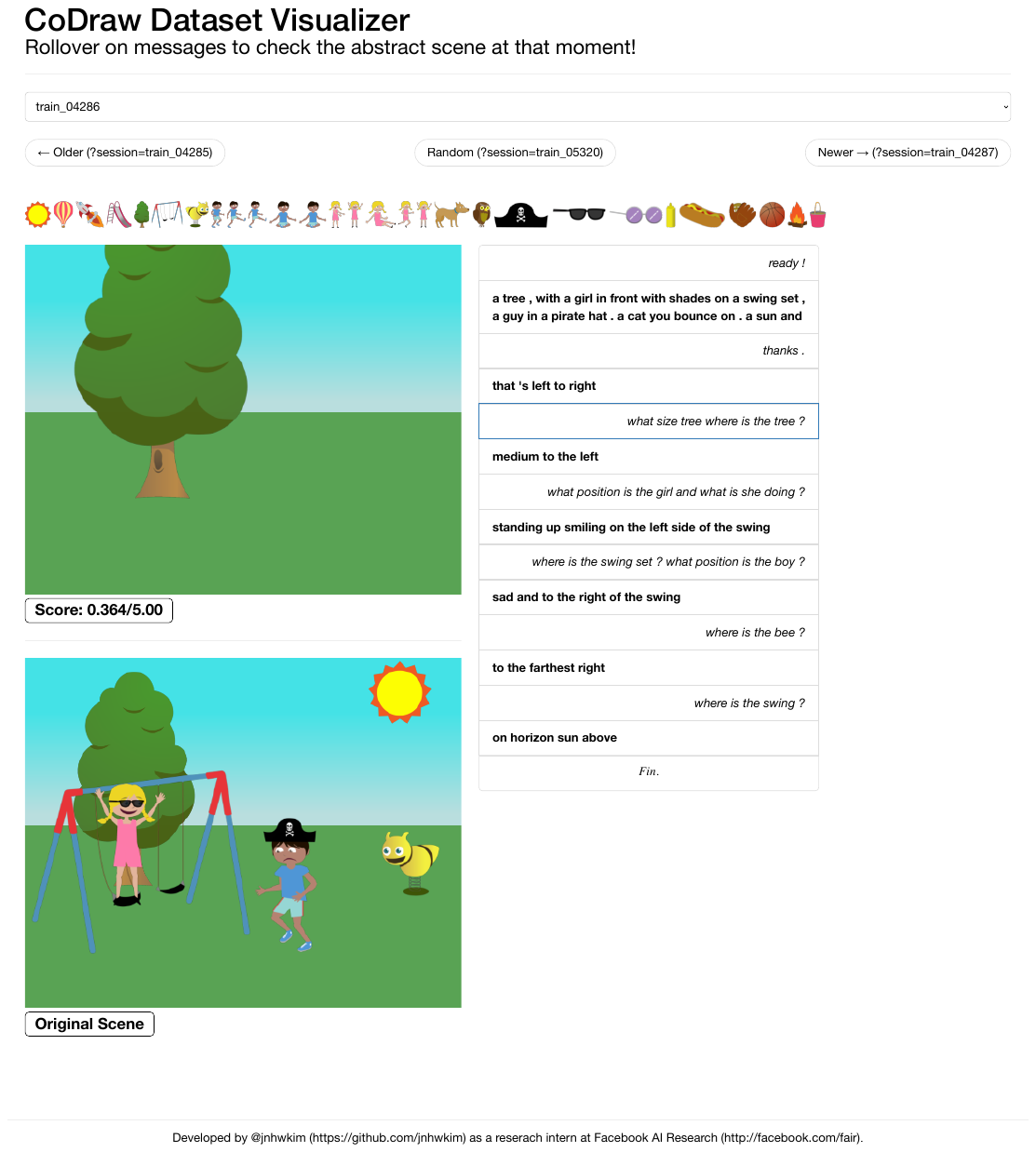}
    \caption{The instruction giver provides underspecified instructions at the first turn. Instead of taking all actions immediately, the instruction follower does many rounds of clarification. From: CoDraw dialogue game 4286, \href{https://creativecommons.org/licenses/by-nc/4.0/}{CC BY-NC 4.0}, scene from \citet{zitnick2013bringing}.}
    \label{fig:codraw-examples-4}
\end{figure*}

Figures \ref{fig:codraw-examples-1}-\ref{fig:codraw-examples-4} exemplify strategies of crowdworkers, showing various levels of commitment to playing the game well. The images are generated with the CoDraw Dataset Visualizer, developed by @jnhwkim at \url{https://github.com/facebookresearch/CoDraw}. Scenes at the top are the state of the reconstructions at the highlighted turns.

\end{document}